\pdfoutput=1
\documentclass[11pt]{article}

\usepackage{acl}
\usepackage{times}
\usepackage{latexsym}
\usepackage{caption}
\usepackage{graphicx}
\usepackage{subcaption}
\usepackage{csquotes}
\usepackage{amsmath, mathtools, multirow}
\usepackage{url}
\usepackage{tabularx}

\usepackage[T1]{fontenc}
\usepackage[utf8]{inputenc}

\usepackage{microtype}

\newcommand{\mn}[1]{{\textcolor{black}{#1}}{\bf}}

\title{Does Simultaneous Speech Translation need Simultaneous Models?}

\author{Sara Papi\textsuperscript{$\clubsuit$,$\diamondsuit$}, Marco Gaido\textsuperscript{$\clubsuit$,$\diamondsuit$}, Matteo Negri\textsuperscript{$\clubsuit$}, Marco Turchi\textsuperscript{$\spadesuit$*} \\
  \textsuperscript{$\clubsuit$}Fondazione Bruno Kessler \\
  \textsuperscript{$\diamondsuit$}University of Trento \\
  \textsuperscript{$\spadesuit$}Zoom Video Communications \\
  \texttt{\{spapi, mgaido, negri\}@fbk.eu, marco.turchi@zoom.us}}

\begin{document}
\maketitle
\begin{abstract}
\newcommand\blfootnote[1]{%
  \begingroup
  \renewcommand\thefootnote{}\footnote{#1}%
  \addtocounter{footnote}{-1}%
  \endgroup
}
\blfootnote{* Work done when working at FBK.}
In simultaneous speech translation (SimulST), finding the best trade-off between high output quality and low latency is a challenging task. To meet the latency constraints posed by different application scenarios, multiple dedicated SimulST models are usually trained and maintained, generating high computational costs. In this paper, also motivated by the increased sensitivity towards sustainable AI, we investigate whether a \textit{single model} trained offline can serve both offline and simultaneous applications under different latency regimes without additional training or adaptation. Experiments on en$\rightarrow$\{de, es\} show that, aside from facilitating the adoption of well-established offline architectures and training strategies without affecting latency, the offline solution achieves similar or better quality compared to the standard SimulST training protocol, also being competitive with the state-of-the-art system.
\end{abstract}

\section{Introduction}

Many application 
contexts,
%\mn{scenarios}
such as conferences and  lectures, 
%and guided tours, 
require automatic speech translation (ST) to be performed in real-time.
%Simultaneous ST (SimulST) systems pursue not only high translation quality but also low latency, that is the elapsed time between the speaker's utterance of a word and the generation of its translation in the target language.
To meet this requirement, Simultaneous ST (SimulST) systems strive not only for high output quality but also for low latency (i.e. the elapsed time between the speaker's utterance of a word and the generation of its translation in the target language).
Balancing quality and latency
is extremely complex as the two objectives are  
%conflicting: in general, especially when the target language has a different word order than the source language, the more a system waits -- higher latency -- the better it translates thanks to a larger context to rely on.
conflicting: in general, the more a system waits -- which implies higher latency -- the better it translates thanks to a larger context to rely on.

% In a simultaneous model, the quality-latency balancing is controlled by a decision policy: the rule that 
SimulST models manage the quality-latency trade-off by means of a decision policy: the rule that determines whether a system has to wait for more input or to emit one or more target words.
The most popular decision policy is the \emph{wait-k}, a straightforward heuristic that prescribes to wait for a predefined number of words before starting to generate the translation. 
%It was initially proposed by \citet{ma-etal-2020-simulmt} for simultaneous machine translation (SimulMT) and then 
Initially proposed by \citet{ma-etal-2020-simulmt} for simultaneous machine translation (SimulMT), the \emph{wait-k} 
%has been
is now
widely adopted in SimulST \citep{ma-etal-2020-simulmt,ren-etal-2020-simulspeech,han-etal-2020-end,chen-etal-2021-direct,zeng-etal-2021-realtrans,ma2021streaming} thanks to its simplicity.
Apart from \emph{wait-k}, other attempts have been made to develop decision policies learned by the SimulST system itself \citep{ma2019monotonic,zaidi2021decision,liu-etal-2021-ustc,liu-etal-2021-cross}, all resulting in 
%not very widespread and
computationally expensive models with limited diffusion.

Regardless of the decision 
%policy and model adopted, a SimulST system is usually trained
policy, SimulST systems are usually trained simulating the conditions faced at inference time,
%\mg{i.e.,}
%, 
%%i.e., only partial input is 
that is
with only a partial input
available \citep{ren-etal-2020-simulspeech,ma-etal-2020-simulmt,han-etal-2020-end,zeng-etal-2021-realtrans,ma2021streaming,zaidi2021decision,liu-etal-2021-ustc}.
Since the size of the partial input -- and consequently of the context that the SimulST system can exploit to translate -- varies according to the latency requirements imposed by real-world applications,\footnote{For instance, the IWSLT 
%competition
SimulST shared task
defines three latency regimes \citep{iwslt_2021} -- $1s$, $2s$, and $4s$ -- and limits of acceptability have been set between $2s$ and $6s$ for the \textit{ear-voice span} depending on different conditions and language pairs \citep{sane-2000,ear-voice-span}.} several models must be trained and maintained to accommodate different quality-latency trade-offs.
This 
%leads to high computational costs which are 
results in high computational costs that contrast
%in contrast with the recent rise of
with rising
awareness on the need to reduce energy consumption \citep{strubell-etal-2019-energy} 
%and artificial intelligence unsustainable development
towards more sustainable AI
\citep{vinuesa2020role,greenai}.
So far, the
benefits of
%training a system 
training systems
on partial inputs have been taken
for 
%granted so far 
granted
and, although
%in the literature there are works employing models trained in offline 
works employing models trained in offline mode are documented in literature \cite{nguyen2021empirical,ma2021streaming},
% \mg{previous works \cite{nguyen2021empirical,ma2021streaming} employed models trained in offline mode for simultaneous inference, such a choice has been motivated by the limited computational power available without questioning}
the 
%actual need for simultaneous training of SimulST systems has never been demonstrated.
indispensability of
simultaneous training in SimulST
has never been
demonstrated.
%Being aware of the social and environmental impact caused by the high computational costs of training SimulST models \citep{greenai},
With an eye at the 
%environmental impact and the
burden 
and environmental impact
of training multiple dedicated models 
%(
for different tasks -- offline, simultaneous -- and latency regimes,
%),
%and its inherent environmental impact,}
in this work we 
%question this approach and ask:
address the following question:
\emph{Does simultaneous speech translation actually need models trained in simultaneous mode?}
% To find an answer to this question, 
% %in this paper 
% we conduct a study aimed at understanding the possible impact of using a model trained offline on real-time performance.
% To find an answer, we study the possible impact of using a model trained offline on real-time performance.
To this end, we 
%explore the development of 
experiment with
a single, easy-to-maintain offline model, which can effectively serve both the simultaneous and offline 
%tasks avoiding any additional training.
tasks.
Specifically, we 
%discuss
explore
the application of 
the widely adopted \emph{wait-k}
%\mn{the \emph{wait-k}}
policy to the offline-trained ST system only at inference time, 
%without the need for 
bypassing
any additional training neither to adapt the model to the simultaneous scenario nor to accommodate different latency requirements.\footnote{Code available at \url{https://github.com/hlt-mt/FBK-fairseq}.}
Through
%through
experiments on two language directions (en$\rightarrow$\{de, es\}), having respectively different and similar word 
%ordering, in this paper we show that:
ordering with respect to the source, we show that:
%\footnote{Dealing with languages that have different word ordering already complicates the translation in offline but even more in simultaneous, since only a partial audio input is at disposal. }
%one having different word ordering (en$\rightarrow$de) while the other having the same word ordering (en$\rightarrow$es),
%We select the English$\rightarrow$German (hereinafter: En-De) and English$\rightarrow$Spanish (hereinafter: En-Es) language pairs, since the first represents a translation from a language with Subject-Verb-Object order (SVO) to a language with Subject-Object-Verb order (SOV) while the second from an SVO to SVO language.
%Dealing with languages that have different word ordering already complicates the translation in offline but even more in simultaneous, since we have only a partial audio input at disposal. 
\begin{itemize}
    
    % \item \mn{Offline training  is a 
    % viable
    % %valid 
    % alternative to the standard %resource-demanding 
    % computation-intensive
    % protocols used for SimulST: in our evaluation setting, it yields a reduction in carbon emission and electricity consumption by a factor of 8 (Section \ref{sec:exp-setting}).}
    
    % \item \mn{Offline training is a viable and less computation-demanding 
    % alternative to current protocols: in our evaluation setting, it yields a reduction in carbon emission and electricity consumption by a factor of 8 (Section \ref{sec:exp-setting}).}
    
    \item In terms of sustainability, offline training  yields considerable reductions -- by a factor of 9 in our evaluation setting -- in carbon emission and electricity consumption (Section \ref{sec:exp-setting}).
    
    \item 
    %Regardless of the word detection strategy adopted, the
    % The offline-trained model outperforms or is at least on par with the model trained in simultaneous for each value of $k$ (Section \ref{sec:offline});
    The offline-trained model outperforms or is
    %at least 
    on par with those trained in simultaneous 
    %for each value of $k$ (Section \ref{sec:offline-results});
    within the \emph{wait-k} policy framework 
    %(Section \ref{sec:offline-results}) \sara{and reduces the carbon emission and electricity consumption by a factor of 8} \mn{[Penso che carbon emission BLA meriterebbe un punto a parte, con riferimento a sez. 4]};
    (Section \ref{sec:offline-results});

\item Recent advancements in offline architectures and training strategies 
    %bring benefit to the quality of offline-trained SimulST systems
further improve output quality
without affecting latency (Section \ref{sec:improve});
%
%
%
% \item The efficiency advantages of offline-trained models also emerge in comparison with the state of the art \citep{liu-etal-2021-cross}:  except for the lowest latency regime, our system is superior in the $2s$-$4s$ latency interval (ear-voice span) with gains up to 4 BLEU (Section \ref{sec:caat}).
\item The effectiveness of offline training also emerges in comparison with the state of the art in SimulST \citep{liu-etal-2021-cross}: except for the lowest latency regime, our system is superior in the $2s$-$4s$ latency interval (ear-voice span) with gains up to 4.0 BLEU
%\mn{points} 
(Section \ref{sec:caat}).
    % for En$\rightarrow$\{De, Es\} language pairs
    %on the more realistic computational-aware metrics
% (Section \ref{sec:caat}).

    %even if, at lower latency, our system performs slightly worse, when the latency increases -- but remains in acceptable intervals --, it gets progressively better results, with gains of up to 4 BLEU points (Section \ref{sec:caat}).
    % All these advantages emerge in the comparison with the current -- but computationally expensive -- state-of-the-art with learned decision policy \citep{liu-etal-2021-cross}: if at lower latency it performs slightly better, when the latency increases -- but remains in acceptable intervals -- our offline-trained model performs increasingly better, with gains up to 4 BLEU points.
    %The superiority of the current state-of-the-art with learned decision policy \citep{liu-etal-2021-cross} is not confirmed against our offline-trained model when the more reliable real-time latency is considered in the systems evaluation (Section \ref{sec:caat}).
    %the current -- but computationally expensive -- state-of-the-art with learned decision policy \citep{liu-etal-2021-cross} obtains impressive scores on ideal latency metrics, but its superiority is not confirmed in practice when considering the real-time latency, in which the computation time matters (Section \ref{sec:caat}).
\end{itemize}

\section{Background}
\label{sec:waitk}
\subsection{\emph{wait-k}}
\label{subsec:k}
The \emph{wait-k} policy requires to wait for a predefined number of words before starting to translate. 
For instance, a system using a wait-3 policy generates the 1\textsuperscript{st} target word when it receives the 4\textsuperscript{th} source word, the 2\textsuperscript{nd} target word when it receives the 5\textsuperscript{th} source word, and so on.
The number of words to wait is controlled by 
%a parameter named $k$.
the $k$ parameter.
%The SimulST 
SimulST
systems based on the \emph{wait-k} policy are usually trained considering the same $k$ used for testing \citep{ren-etal-2020-simulspeech,ma-etal-2020-simulmt,zeng-etal-2021-realtrans}
while, in theory, its value can 
be different
%\mn{differ}
%.
%However, this parameter may be different in theory
between the training and testing 
%phases: a
phases. A
parameter $k_{train}$ can indeed be used to mask words at training time, 
%and
while a parameter
% %\mg{different value} 
% $k_{test}$ 
% %can be used 
% %%at inference time 
% %to directly influence
% %\mg{allows for controlling} 
% \sara{allows 
% %to directly control
% \mg{for directly controlling}}
% the latency of the system at inference time according to the
% %speed 
% requirements posed by the target application scenario.
$k_{test}$ can be used to directly control the latency of the system at inference time according to the requirements posed by the target application scenario.

%As 
Since many values of $k_{train}$ can be used to train the SimulST systems, even for identical values of $k_{test}$, the standard approach involves performing
several trainings to obtain the best translation quality while satisfying different latency requirements.  
In SimulMT, \citet{Elbayad2020} tried to 
%solve this problem 
%overcome this large amount 
%bypass
\mn{avoid}
this large number
of experiments by 
%\mg{exposing the model with different values of $k_{train}$, which is sampled at each iteration.}
exposing the model to different values of $k_{train}$ sampled at each iteration.
%developing a method that explore
%%to support \mn{[IN CHE SENSO ``support''?]} 
%multiple values of $k_{train}$ on a single system during 
%%training but, surprisingly, 
%training. 
Surprisingly,
they achieve the best performance on several $k_{test}$ using a single value of $k$ for training 
($k_{train}=7$).
However, it is not clear if such a rule applies to SimulST, leaving the problem of performing 
a large number of 
%\mn{many}
trainings still unsolved.

%In addition, it is not known in advance which value of $k$ corresponds to a certain latency, thus requiring a SimulST model (trained for a specific value of $k_{train}$) to be tested on multiple $k_{test}$ to identify the actual latency reached.

\subsection{Word detection for \emph{wait-k} in SimulST}
\label{subsec:worddet}
Since 
%in 
SimulMT 
%the input is 
operates on
%\mn{[``operates on'' ERA PER EVITARE DI RIPETERE INPUT 2 VOLTE NELLA STESSA FRASE. SE NON VA BENE, BISOGNA RIFORMULARE IN QUALCHE MODO.]}
a stream of words, applying the \emph{wait-k} is straightforward because 
%the information about the number
the number
of received words is explicit
in the input. Conversely, its application to SimulST is 
%not so easy because 
complicated by the fact that
the input is an audio stream and the number of received words has to be inferred by means of a so-called \textit{word detection} strategy.
%To detect how many words are contained in the audio, a word detection strategy is used, representing the method by which these words are counted. 

Two main categories of word detection strategies are currently employed by the community: fixed \citep{ma-etal-2020-simulmt}, and adaptive \citep{ma-etal-2020-simulmt,ren-etal-2020-simulspeech,zeng-etal-2021-realtrans,chen-etal-2021-direct}.
The fixed strategy is the easiest 
%way to address the problem, 
approach,
as it assumes that a fixed amount of time is required to pronounce every word disregarding the information actually contained in the audio.
% problem: disregarding the information contained in the audio, it assumes that a fixed amount of time is required to pronounce every word.
%This amount of time is usually determined by computing the average word duration in the dataset \citep{ma-etal-2020-simulmt}. 
%The resulting value is used as a time-step for an external module that decides if the system has to wait for more source input or emit a translated word.
%On the contrary, the adaptive word detection strategy 
In contrast, adaptive word detection determines the number of uttered words by looking at the content of the audio.
%The decision about the number of words can be taken through 
%\mn{The number of uttered words can be either determined  by means}
This can be done either by means of
an Automatic Speech Recognition (ASR) decoder \citep{chen-etal-2021-direct},\footnote{This solution involves the use of two separate synchronized decoders (one for simultaneous ASR and one for ST) and will not be analyzed in this work due to the higher computational costs of
training 
%\mn{setting up}
a double decoder architecture. 
%\mg{that also increase the computational-aware latency}.
} 
%or through
%\mn{or inferred by means of}
or by means of a Connectionist Temporal Classification \citep{ctc-2006} -- 
%or 
CTC -- module \citep{ren-etal-2020-simulspeech,zeng-etal-2021-realtrans}, every time a speech chunk is received by the system.

%is taken by a module that is usually based on Connectionist Temporal Classification -- or CTC -- \citep{ctc-2006}, which is responsible for directly detecting the number of words every time a speech chunk is received by the system \citep{ren-etal-2020-simulspeech,zeng-etal-2021-realtrans}.
%In this way, the word detection is adapted to the effective content of the audio.}
%at the cost of employing a more complex mechanism.}

%On one hand, the fixed strategy represents the simplest technique to apply to a SimulST system. On the other hand, it 
In its simplicity, the fixed strategy does not consider various aspects of the input speech, such as different speech rates, duration, pauses, and silences.
%Thus, even 
%This module is usually based on Connectionist Temporal Classification -- or CTC -- \citep{ctc-2006} that predicts the number of words in each speech chunk, 
% %are based on the number of word predicted via Connectionist Temporal Classification -- or CTC -- \citep{ctc-2006}, 
%so that the word segmentation is adapted to the effective content of the audio. 
%Building a system that starts to output part of the translation when a word is actually present in the source speech represents a valid alternative to the fixed strategy that can improve the performance of the SimulST system. 
For instance, if there are no words in the speech (e.g. in the case of pauses or silences), the fixed strategy forces the system to output something even if it cannot rely on sufficient context.
% In the opposite case, in which more than one word is pronounced in a speech chunk, the adaptive strategy guides the model to output more than one target words while the fixed one  forces the emission of only one word, consequently accumulating a delay. 
% Although the adaptive strategy seems to be more faithful to the actual audio phenomena, conflicting results are  reported in literature, some in support of the adaptive strategy \citep{zeng-etal-2021-realtrans} while others showing no advantage from its  application \citep{ma-etal-2020-simulmt}.
%
%
%
% \mn{In the opposite case, in which more than one word is pronounced in a speech chunk, the fixed strategy  forces the emission of only one word -- consequently accumulating a delay -- while the adaptive one guides the model to output more than one target words.}
In the opposite case, in which more than one word is pronounced in a speech chunk, the fixed strategy  forces the emission of only one word, consequently accumulating a delay.
By trying to guess the actual number of words contained in a speech chunk, the adaptive strategy is in principle more faithful to these audio phenomena.
However, conflicting results are reported in literature, some in support of the adaptive strategy \citep{zeng-etal-2021-realtrans} while others showing no advantage from its application \citep{ma-etal-2020-simulmt}.

%\section{Offline Training for SimulST}
\section{Do we need Simultaneous training?}
%\mn{[GIA' IL TITOLO DEL PAPER E' UNA DOMANDA; RIESCI A METTERE QUI UN TITOLO CHE NON LO SIA? tipo ``Offline training for SimulST'']}
\label{sec:offline}
While at training time the SimulST system has the entire audio available, at inference time it receives a partial, incremental input.
This mismatch between offline training and simultaneous testing makes the system vulnerable to 
%exposure bias, whose
exposure bias \cite{Ranzato15seq}. To mitigate this potential problem, SimulST models are trained
%, whose impact is usually mitigated by performing training 
under simulated simultaneous conditions. 
%%This simultaneous training is
%This is realized on an attentive model 
On an attentive model, this simultaneous training is realized
by masking future audio frames when computing the encoder-decoder attention. 
For a \emph{wait-k} SimulST system, the choice of 
%which audio frames have 
the audio frames
to be masked depends on two factors: the value of $k_{train}$ and the word detection strategy.
The $k_{train}$ value determines the number of source words to mask (e.g., in the case of wait-3, the
%\mn{1\textsuperscript{st}}
first
target word 
%can look only from the 1\textsuperscript{st} to the 3\textsuperscript{rd} 
is generated by looking at the first three
source words and so on). 
The word detection strategy identifies the source words from the audio by detecting the number of frames each one corresponds to. Thus, the encoder-decoder attention is computed by 
limiting
%\mn{forcing}
each target word to only attend to the audio frames that correspond to the previous $k_{train}$ source words 
%according to 
identified by
the word detection strategy.
%determined by the word detection strategy, corresponding to the previous $k$ source words determined by $k_{train}$. 
As a result, testing different word detection strategies requires training several systems, which in turn are trained with different values of $k_{train}$ to obtain different latencies.

In this paper, we question the need 
%for this large amount of 
of all these
experiments by investigating whether the simultaneous training of the ST systems 
%is actually required to perform the simultaneous task.
%actually brings benefits in performing the SimulST task and if it 
is indispensable to obtain a good quality-latency trade-off.
Within the framework of the \emph{wait-k} policy, we
explore
%\mg{assess} 
the ability to translate in real-time 
of
an offline-trained system that is neither trained nor adapted to the simultaneous scenario.
%\footnote{Although been hypothesized \citep{nguyen2021empirical}, no work is present in the literature investigating the potential of the offline systems to serve simultaneous tasks.}
%To translate in real-time, we force the offline-trained system to completely recompute the encoder-decoder attention every time a speech segment is received.
To obtain a simultaneous prediction from the offline system, we add a \textit{pre-decision module} after the encoder at inference time.
%\mt{qui spezzerei la frase. Ripartendo con``It incorporates''}
% ,
% %.
% %only at inference time.
% %Its 
% whose
% role is to 
% %\mg{assist the \emph{wait-k} policy (with a given $k_{test}$) in the decision}
% decide whether 
% to wait or to emit words each time a new speech chunk is received
% by incorporating the logic of the word detection strategy and
% according to the selected $k_{test}$.
Its role is to incorporate 
%\sara{that incorporates}
the logic of the word detection strategy 
%\mn{[NON SI CAPISCE COSA VUOI DIRE CON ``that incorporates the logic of the word detection strategy'']} 
to decide whether to wait or to emit words
%each time 
when
a new speech chunk is received, according to the selected $k_{test}$.
In particular, 
it
%\mn{the pre-decision module}
takes as input the encoder states representing the received audio chunk and applies the word detection strategy (either fixed or adaptive) to obtain the number of source words present in the input. If this number is equal or exceeds $k_{test}$, the 
%pre-decision 
module 
%enables
activates
the decoding part of the model and a word is emitted, otherwise it keeps reading the source speech.
%The \emph{wait-k} with fixed strategy is realized by waiting for a fixed amount of time, which is $k_{test}$ times the dimension of the speech segment, before starting the generation of the translation. The \emph{wait-k} with adaptive strategy is realized by waiting until the number of words detected from the audio source, which is updated every time a new speech segment is received, exceeds the value of $k_{test}$.

Since the offline system is not trained for the simultaneous task, 
%this choice
the choice of $k_{test}$ and word detection strategy
% %at inference time 
% for testing
%is 
are
not constrained to 
%the parameters 
those
used during training as in the native SimulST case.
%can be made freely without completely re-training the model, unlike the native SimulST case.
%where they depend on the parameters used during training.
Indeed, an offline model is trained by always attending to the entire source input. 
Different from
%\mg{Unlike} 
the simultaneous training mode, the encoder-decoder attention is computed without masking, that is by considering past, current, and future information.
Although this avoids multiple training for each $k_{train}$ and word detection strategy, it also exposes the model to operate in conditions different from its training setup, as it is not used to receive partial inputs.
%This makes the offline model not used to receive partial incremental inputs. 
%As a consequence, although it avoids  multiple training for each $k_{train}$ and word detection strategy, 
%%the  offline-trained system is not
%the  offline training mode might not seem theoretically suitable to perform the simultaneous task.
%
%
%
%
% To verify whether 
% \sara{the} exposure bias 
% \sara{given by this mismatch in training and testing conditions} constitutes or not a real limitation for the offline-trained system, we conduct a systematic study by analyzing its quality-latency performance in the simultaneous scenario. 
To 
%verify whether  the exposure bias given by this mismatch in training and testing conditions constitutes or not a real limitation,
check if the exposure bias given by this mismatch in training and testing conditions constitutes a real limitation,
we conduct a systematic analysis of the quality-latency performance of the offline-trained system in the simultaneous scenario. 
%In the following, we present the results of this study by comparing
To this aim, we compare
the offline-trained system with the same model trained in simultaneous mode by varying the value of $k_{train}$ and the word detection strategy.

\section{Experimental Settings}
\label{sec:exp-setting}
We perform all our experiments on
%For all our experiments, we use 
the en$\rightarrow$\{de, es\} sections of the MuST-C dataset \citep{CATTONI2021101155}. All the results 
%present in the paper 
presented
are given on the corpus test set (tst-COMMON).
We use the Transformer architecture \citep{transformer}
%The encoder-decoder architecture used for the experiments consists of a Transformer \citep{transformer} model 
with the integration of the CTC in the encoder \citep{liu2020bridging,gaido-etal-2021-ctc}, which is used to realize the adaptive word detection strategy.
% but also to 
% %apply the CTC-compression mechanism to 
% compress
% the input with the aim of speeding up the simultaneous inference (more details 
% %are provided 
% in Appendix \ref{sec:CTC-compress}). 
The hyper-parameters, training and inference details are presented in Appendix \ref{subsec:transformer}.

For the evaluation, we adopt 
%the popular and commonly used metrics for SimulST:
BLEU\footnote{BLEU+case.mixed+smooth.exp+tok.13a+version.1.5.1} \citep{post-2018-call} for quality, and 
Length Adaptive Average Lagging \citep{papi-etal-2022-generation} -- or LAAL -- for latency, which is the modified version of the popular Average Lagging for speech \citep{ma-etal-2020-simulmt} that correctly evaluates both shorter and longer predictions with respect to the reference. We report the simultaneous results in LAAL-BLEU graphs where each curve corresponds to a system trained using a different value of $k_{train}$ and each point to a different $k_{test}$. 
The set of $k$ values used for both 
training the simultaneous model and testing all the 
models is $k=\{3,5,7,9,11\}$.
We also report 
%in the graphs 
the results of the offline generation using the greedy search
%, which emits the most probable token and is also used in our simultaneous setting because is faster, 
and 
%of 
the beam search with 
% $beam\_size=5$, commonly used in offline ST. 
the $beam\_size=5$ commonly used in offline ST.

\begin{figure}[tb]
     \centering
     \begin{subfigure}[b]{0.475\textwidth}
         \centering
         \includegraphics[width=\textwidth]{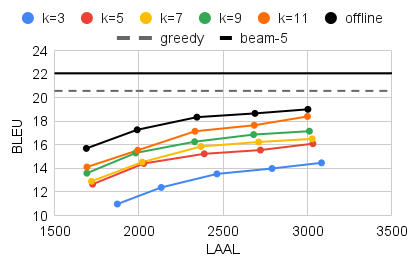}
         \caption{English $\rightarrow$ German}
     \end{subfigure}
     \hfill
     \begin{subfigure}[b]{0.475\textwidth}
         \centering
         \includegraphics[width=\textwidth]{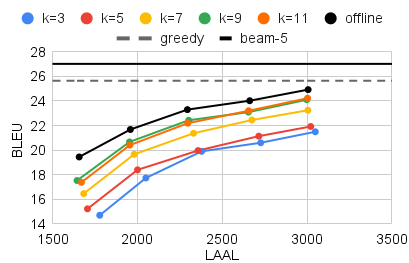}
         \caption{English $\rightarrow$ Spanish}
     \end{subfigure}
        \caption{LAAL-BLEU curves of \emph{wait-k} with fixed word detection strategy.}
        \label{fig:fixed}
\end{figure}

%%% XXXX
\begin{figure}[tb]
     \centering
     \begin{subfigure}[b]{0.475\textwidth}
         \centering
         \includegraphics[width=\textwidth]{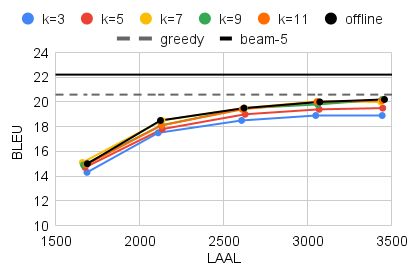}
         \caption{English $\rightarrow$ German}
     \end{subfigure}
     \hfill
     \begin{subfigure}[b]{0.475\textwidth}
         \centering
         \includegraphics[width=\textwidth]{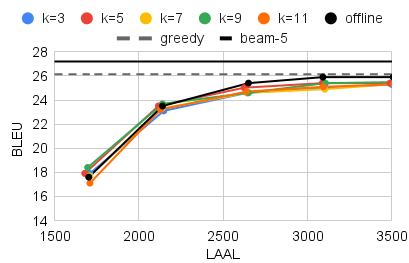}
         \caption{English $\rightarrow$ Spanish}
     \end{subfigure}
        \caption{LAAL-BLEU curves of \emph{wait-k} with adaptive word detection strategy.}
        \label{fig:adaptive}
\end{figure}

\paragraph{Carbon Footprint.}
Each training 
%contributes
contributed
an estimate of 70.3 kg of $\text{CO}_{2eq}$ to the atmosphere and used 184.7 kWh of electricity. This assumes 116 hours of runtime, a carbon intensity of 380.539g $\text{CO}_{2eq}$ per kWh, 4 NVIDIA Tesla K80 GPUs (utilization 93\%), and an Intel Xeon CPU E5-2683 v4 (utilization 100\%).\footnote{The social cost of carbon uses models from \citep{ricke2018country} and 
%this statement \mn{[Cosa vuoi dire con ``this statement''? Qualcosa tipo ``and our carbon emission estimates have been obtained using the...''?]} and 
carbon emissions information was
%generated 
estimated
using the \emph{experiment-impact-tracker} \citep{henderson2020systematic}.}
This means that training a single offline model instead of a model for each value of $k_{train}$ (in our case, 5 models) and for each word detection strategy (in our case, 2 strategies) allows us to save 
%$(5-1) \cdot 2 = 8$
$5 \cdot 2 - 1 = 9$
experiments, 
%which corresponds to 562 kg of $\text{CO}_{2eq}$ and 1477.6 kWh
amounting to 632.7 kg of $\text{CO}_{2eq}$ and 1662.3 kWh 
of electricity for each language. 
%\mn{[Mi piace questa parte. Penserei di riprenderne bene il messaggio in Intro e conclusioni. Poi non so se ci sarebbe altro da dire e che possiamo mettere in appendice.]}

\section{Results}
\label{sec:offline-results}

\paragraph{Fixed Word Detection.}
The results of the \emph{wait-k} models with fixed word detection are shown in Figure \ref{fig:fixed}.
The LAAL-BLEU curves 
%show
indicate
that the latency of all the systems lies between $1700ms$ and $3000ms$, staying in a medium-high latency regime\footnote{
%Throughout the paper, we talk about \textit{low} latency regime when the delay between the time in which the speech is heard and the output translation is received is under $1000ms$, \textit{medium} when this time is between $1000$ and $2000ms$, and \textit{high} when it is over $2000ms$ \citep{iwslt_2021}.
Henceforth referring to \citep{iwslt_2021}, we 
%isolate
consider
three latency regimes depending on the delay $d$ between the time in which the speech is heard and the output translation is received. These are: \textit{low} when $d$<$1000ms$, \textit{medium} when $1000$<$d$<$2000ms$, and \textit{high} when $d$>$2000ms$.}
for both language pairs.
%The
Translation
quality is lower for 
%en-de and ranges 
en-de, for which it ranges
from 11 to 19 BLEU, while for en-es it ranges from 14 to 25 BLEU. The difference in performance between the two language pairs is coherent with the results of the offline generations (both greedy and beam-5) and justified by the different level of difficulty 
%posed by the two target languages
when translating into the two target languages (having respectively similar and different word ordering with respect to English).
%and derive from the greater ease of an ST system to translate between languages having the same word ordering.
The curves of the simultaneous-trained systems also show a tendency: if 
%the 
$k_{train}$ increases, both the quality and latency 
%improve.
improve 
(e.g. on en-de, the $k$=11 curve lies higher -- indicating better quality -- and more leftward -- lower latency -- than the others).
%This phenomenon is particularly evident when we consider the offline-trained model (in solid black) since it outperforms, at every latency regime, the systems trained in simultaneous, with gains from 1 to 7 BLEU for en-de and from 1 to 6 BLEU for en-es. 
Interestingly, the 
%offline-trained model (in solid black) outperforms 
offline-trained models (in solid black) outperform
the systems trained in simultaneous at every latency regime, with gains from 1 to 7 BLEU
%\mn{points} 
for en-de and from 1 to 6 BLEU
%\mn{points} 
for en-es.
%It means that,
This indicates that,
to achieve the best performance and independently from the $k_{test}$
%value 
used, the offline-trained model represents the best choice, at least 
%for a SimulST system adopting 
for the fixed
%word detection 
strategy. 

%%% XXXX
% \begin{figure}[htb]
%      \centering
%      \begin{subfigure}[b]{0.46\textwidth}
%          \centering
%          \includegraphics[width=\textwidth]{images/segmentation/german_adaptive.png}
%          \caption{English $\rightarrow$ German}
%      \end{subfigure}
%      \hfill
%      \begin{subfigure}[b]{0.46\textwidth}
%          \centering
%          \includegraphics[width=\textwidth]{images/segmentation/spanish_adaptive.png}
%          \caption{English $\rightarrow$ Spanish}
%      \end{subfigure}
%         \caption{AL-BLEU curves of \emph{wait-k} with adaptive word detection strategy.}
%         \label{fig:adaptive}
% \end{figure}

\paragraph{Adaptive Word Detection.}
The results of the \emph{wait-k} models with adaptive word detection are shown in Figure \ref{fig:adaptive}.
%As 
%%for
%with
%the fixed strategy, 
%the systems lie
%systems latency lies
The systems latency lies
between $1700ms$ and $3500ms$ and, as with the fixed strategy, the quality is higher for en-es (from 15 to 26 BLEU) than for en-de (from 14 to 20 BLEU).
Looking at 
%both Figure
Figures
\ref{fig:fixed} and \ref{fig:adaptive}, we observe that the overall translation quality yielded by the adaptive strategy is higher compared to that of the fixed one.  Moreover, the fixed strategy curves are far from being comparable with their offline greedy values 
%(dashed straight lines),
(dashed lines),
while the adaptive strategy curves almost reach them at higher latency. 
However, the models with fixed word detection perform better at lower latency, with a gain of 1 BLEU %\mn{point} 
for en-de and 2 BLEU
% \mn{points} 
for en-es.
%These results highlight that, at least for the \emph{wait-k} policy, there is not a clear winner between fixed and adaptive strategy. Instead, the choice of the word detection should be done according to other factors as the desired latency, which depends on the particular use case.
%These results highlight that 
In light of these results,
there is not a clear winner between 
%fixed and adaptive strategy.
the two word detection 
strategies.
%\mn{strategies, ...}
%and that the choice of the word detection should be done instead according to other factors as the desired latency, which depends on the particular use case.
From Figure \ref{fig:adaptive}, we 
%can also 
also
notice that
%Looking again at Figure \ref{fig:adaptive}, we can notice another interesting trend: 
the adaptive curves are very close to each other, in contrast with the fixed case.
This phenomenon indicates that, in the case of the adaptive strategy, changing
%the $k_{train}$
$k_{train}$ does not significantly influence
the 
model performance.
%\mn{performance of a model trained in simultaneous mode.}
This 
%means
suggests that the offline-trained model (comparable to a model trained with $k_{train}=\infty$)
%will perform as good as
should
%might
be on par with the simultaneous-trained ones, a consideration
%consideration that is also
corroborated by the 
%behaviour 
trend
of the offline-trained system curves (in solid black) that are always above 
%or nearly on par 
or on par
with those of the simultaneous-trained systems.

All in all, we can conclude that, 
%within the framework of the \emph{wait-k} policy, 
when using the \emph{wait-k} policy,
%there is no need to train the model online for either of the two word detection strategies since 
\textbf{the offline-trained model achieves similar or even better results compared to the same models trained in simultaneous mode}. 
% In the following section, we explore the potential of this finding by adopting the most promising offline architectures and training techniques with the aim of improving the quality-latency balancing of our offline-trained SimulST system.
Based on this finding, in the next section we explore the actual potential of offline training for SimulST
by adopting the most promising offline architectures and training techniques to improve the quality-latency balancing of our systems.

\section{Leveraging Offline Solutions}
\label{sec:improve}
% In Section \ref{sec:offline}, we discovered that the offline-trained model can be used in real-time, achieving similar or better results compared to simultaneous-trained models.
% %, regardless of the word detection strategy adopted. 
% This finding leads to significant advantages in terms of reducing the computational costs of the SimulST systems. 
% First, an offline model can be used in real-time by simply applying fixed or adaptive word detection strategies at inference time, meaning that only one model has to be trained and maintained to serve both offline and simultaneous tasks. 
Offline training brings considerable advantages in terms of reducing the computational costs of  SimulST technology. First, only one model can be trained and maintained to serve both offline and simultaneous tasks without performance degradation.
Second, 
%contrarily with the simultaneous-trained models,
contrary to the simultaneous-training mode,
the choice of the word detection strategy at run-time does not depend on the strategy used during  
% training but can be \mg{adapted} 
training. Rather, it can be made according
to the specific use case, making the offline-trained model more flexible.
This also means that other decision policies can be applied to the offline-trained system without the need to re-train it from scratch.
%Moreover, only one model for both offline and simultaneous ST tasks has to be maintained.

Using a single offline-trained model 
%not only speeds up the development but also opens up the possibility of adopting well-established offline architectures or techniques without performing any additional training to adapt them to the simultaneous scenario.
%does 
not only speeds up its development
%;
%since there is only one model to maintain, 
but also
%, it 
opens up the possibility to directly adopt powerful offline architectures and techniques without performing any additional training nor adaptation to the simultaneous scenario.
%offline system in simultaneous.
%improve the overall performance of the SimulST system.
%In one hand, the use of high performing offline models can be explored. In the other hand, techn
In the following, we test 
% this last hypothesis
this hypothesis
to find out whether 
%the
recent
architectural improvements (Section \ref{subsec:architecture}) and data augmentation techniques (Section \ref{subsec:KD}) 
%usually applied to 
designed for
offline ST also have a positive impact 
%on
in
SimulST.
%by using a popular high performing architecture and a firm-grounded data-augmentation technique in the offline scenario to find if their improvements also impacts the simultaneous task.

%%%XXXX FIG 3
% \begin{figure}[tb]
%      \centering
%      \begin{subfigure}[b]{0.495\textwidth}
%          \centering
%          \includegraphics[width=\textwidth]{images/conformer/german_conformer.png}
%          \caption{English $\rightarrow$ German}
%      \end{subfigure}
%      \hfill
%      \begin{subfigure}[b]{0.495\textwidth}
%          \centering
%          \includegraphics[width=\textwidth]{images/conformer/spanish_conformer.png}
%          \caption{English $\rightarrow$ Spanish}
%      \end{subfigure}
%         \caption{AL-BLEU curves of the Transformer- and Conformer-based architectures.}
%         \label{fig:conformer}
% \end{figure}

% \subsection{Scaling Architecture}
% \label{subsec:architecture}
In 
%the last
recent
years, many architectures have been proposed to address the offline ST task \citep{wang-etal-2020-fairseq,inaguma-etal-2020-espnet,le-etal-2020-dual,papi-etal-2021-speechformer}. Among them, the Conformer
\citep{gulati20_interspeech} has recently shown impressive results both in speech recognition, for which it was initially
proposed, and in speech translation \citep{inaguma2021non}.
The main aspects characterizing this encoder-decoder architecture are related to the encoder part.
Inspired by the Macaron-Net \citep{lu2019understanding}, the Conformer encoder is built with a sandwich structure and integrates the relative sinusoidal positional encoding scheme 
%of the Transformer-XL 
\citep{dai-etal-2019-transformer}.
%While in the original implementation, the decoder part is a Recurrent Neural Network (a single LSTM layer), in its ST version by \citet{inaguma2021non} it is a classic Transformer. 

Given the promising results 
%achieved by this architecture in the offline scenario, we choose to test whether it also impacts or not the quality and latency in simultaneous.
it achieved
in the offline scenario, we choose to test 
if this architecture also brings quality and latency gains in SimulST.
%We present a comparison between Conformer- and  Transformer-based architecture.\footnote{Since we found in Section \ref{sec:offline} that neither of the segmentation strategies works better, we include both in the following analysis.} 
Since we found in Section \ref{sec:offline} that fixed and adaptive word detection strategies have their own use 
%cases, 
cases (their best results are observed at different latency regimes, respectively low for fixed and medium-high for adaptive), 
%they achieve better results respectively at low and medium-high latency
%
%the former giving better results at low latency and the latter)}
%\mn{[COSA VUOI DIRE CON ``have their own use cases''?]},
we compare Conformer- and Transformer-based architectures using both strategies.
For the offline training of 
%the Conformer,
Conformer,
we follow the same procedure used for 
%the Transformer.
Transformer.
%The details of 
Details about
the model hyper-parameters are presented in Appendix \ref{subsec:conformer}.

\begin{figure}[tb]
     \centering
     \begin{subfigure}[b]{0.475\textwidth}
         \centering
         \includegraphics[width=\textwidth]{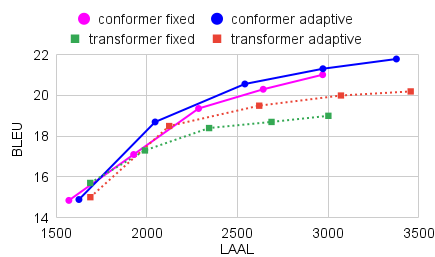}
         \caption{English $\rightarrow$ German}
     \end{subfigure}
     \hfill
     \begin{subfigure}[b]{0.475\textwidth}
         \centering
         \includegraphics[width=\textwidth]{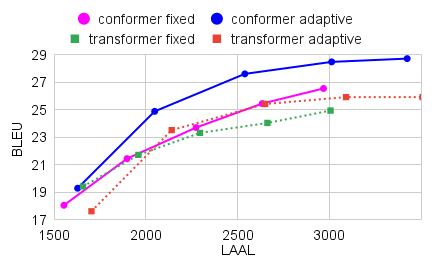}
         \caption{English $\rightarrow$ Spanish}
     \end{subfigure}
        \caption{LAAL-BLEU curves of the Transformer- and Conformer-based architectures.}
        \label{fig:conformer}
\end{figure}

\subsection{Scaling Architecture}
\label{subsec:architecture}

%\paragraph{Results.}
The offline results of both 
%Transformer and Conformer models
architectures
are presented in Table 
%\ref{tab:conformer} and 
\ref{tab:conformer}, while their simultaneous curves are shown in Figure \ref{fig:conformer}.

\begin{table}[htb]
    \centering
    \small
    \setlength{\tabcolsep}{4pt}
    \begin{tabular}{c|cc|cc}
        \multirow{2}{*}{Model} & \multicolumn{2}{c|}{\textbf{En-De}} & \multicolumn{2}{c}{\textbf{En-Es}} \\
        \cline{2-5}
          & \texttt{greedy} & \texttt{beam-5} & \texttt{greedy} & \texttt{beam-5} \\
         \hline
         Transformer & 20.6 & 22.2 & 26.1 & 27.2 \\
         Conformer & 23.3 & 24.8 & 28.5 & 29.6 \\
    \end{tabular}
    \caption{BLEU results of the offline generation.}
    \label{tab:conformer}
\end{table}

As previously noticed by \citet{inaguma2021non}, 
%the 
Conformer outperforms 
%the 
Transformer in 
%the 
offline generation. The improvements,  of at least 2.4 BLEU points, are valid both for greedy and beam search.
From Figure \ref{fig:conformer}, we can see that 
%Conformer also has an overall better performance in simultaneous compared to Transformer. 
Conformer outperforms Transformer also in the simultaneous setting.
% The Conformer with fixed word detection outperforms the Transformer with fixed word detection and also for the adaptive strategy we can observe an equivalent behaviour. In both cases, the BLEU gain is more evident as the latency increases.
This holds both  
%for the fixed and adaptive word detection strategies, with BLEU gains that increase at higher latency regimes.
for fixed and adaptive word detection, with larger BLEU gains at higher latency regimes.
% We can also notice a similar trend between Conformer and Transformer in terms of word detection strategy: the fixed performs better at lower latency while being outperformed by the adaptive as soon as the latency starts to increase. 
As far as word detection strategies are concerned, we 
%can 
also notice a similar trend between Conformer and Transformer: the fixed one performs better or on par at lower latency while being outperformed by the adaptive one when the latency increases.

%As better results are shown by
In light of the better results obtained by
Conformer,  
we 
%can 
conclude that \textbf{improving the architecture of the offline system
%has also
also has a positive impact on its simultaneous performance}, 
enhancing translation quality without affecting latency.
%thus we can build a single model for both tasks without the need of any additional ad-hoc training for simultaneous.
%with all the advantages of making one single training to perform both tasks.

\subsection{Scaling Data}
\label{subsec:KD}
Data augmentation is a common practice used to improve 
%the performance of systems of different types.
systems performance.
%\mn{One way to approach data augmentation is by applying}
%\sara{One way to do} data augmentation is by applying
One approach to data augmentation is to apply knowledge distillation 
%(KD) that 
(KD), which
was introduced to transfer knowledge from big to small models \citep{hinton2015distilling}.
%It can be accomplished using different methods and, among them, the
Among the possible methods,
sequence-level KD \citep{kim2016sequencelevel} is one of the most popular ones in ST thanks to its application simplicity 
%while bringing consistent improvements
and the consistent improvements observed
%in various settings.}
\citep{potapczyk-przybysz-2020-srpols,xu-etal-2021-niutrans,gaido2022Students}.
%The sequence-level
Sequence-level
KD consists of replacing the target 
references
%\mn{elements}
of a given parallel training corpus with the predicted sequences generated by a
%(big) 
teacher model (usually, an MT model), from which we want to distill the 
%\emph{dark knowledge}.
knowledge 
%to be transferred to a
%(smaller) 
to a student model.

\begin{figure}[tb]
     \centering
     \begin{subfigure}[b]{0.475\textwidth}
         \centering
         \includegraphics[width=\textwidth]{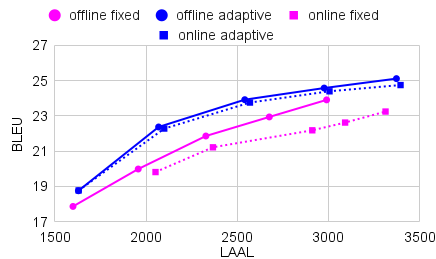}
         \caption{English $\rightarrow$ German}
     \end{subfigure}
     \hfill
     \begin{subfigure}[b]{0.475\textwidth}
         \centering
         \includegraphics[width=\textwidth]{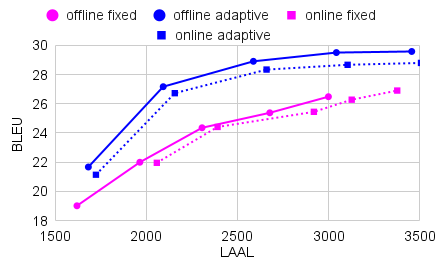}
         \caption{English $\rightarrow$ Spanish}
     \end{subfigure}
        \caption{LAAL-BLEU curves of offline- and simultaneous-trained Conformer models with sequence-level KD.}
        \label{fig:kd}
\end{figure}

To investigate the effects of 
%the data augmentation techniques
such a knowledge transfer
on quality and latency, we apply 
%the 
sequence-level KD to our offline-trained SimulST system.
%To enhance our system performance, we apply the sequence-level KD with the aim of 
%choose to train it using the sequence-level KD to improve both quality and latency in the simultaneous scenario.
To this end, we translate the transcripts present in the en$\rightarrow$\{de, es\} sections of MuST-C with an MT model (more details are provided in Appendix \ref{subsec:mt})
%to obtain the synthetic translations. Then, to build new data, we substitute the gold translations with the MT-generated ones.
and we substitute the gold translations with the MT-generated ones to build new data.
As in \citep{liu-etal-2021-cross}, 
%for training
to train
the models 
%we simply use 
we use
both gold and synthetic data by concatenating them.
Since the performance of the Conformer model scales with data \citep{gaido-etal-2022-efficient} and is better compared to that of Transformer (Section \ref{subsec:architecture}), we adopt the Conformer for the following study.
We extend our analysis to the simultaneous-trained systems to verify if the offline-trained one continues to perform at least on par with them
%. For this reason, we also 
and we
report the best simultaneous-trained system curve for each word detection strategy. 
%To verify that the offline-trained Conformer continues to perform at least on par with its online-trained version, we also report the best online-trained Conformer results considering both the segmentation strategies.

The effects of the additional KD data are shown in Figure \ref{fig:kd}. 
%Comparing Figure \ref{fig:conformer} and \ref{fig:kd}, we can 
Compared to Figure \ref{fig:conformer}, we
notice a performance improvement 
%when using KD that
that
comes without sacrificing latency.
%compared to that of Section \ref{subsec:architecture}:
On en-de, the quality of the offline-trained Conformer with KD ranges from 18 to 25 
%BLEU for en-de, 
BLEU,
against the previous 15 to 22 
%BLEU and 
BLEU. On en-es, it
ranges from 19 to 30 
%BLEU for en-es, 
BLEU,
against the previous 18 to 29 BLEU.
%Hence, the use of the KD technique improves the translation quality by 1 to 3 BLEU without negatively affecting the latency.
Moreover, the offline-trained system 
%(curve in solid) 
(solid curves)
is still better or at least comparable with the simultaneous-trained ones 
%(curves in dotted) 
(dotted curves)
for both language pairs.
From Figure \ref{fig:kd}, we 
%can also
also
notice that 
%the 
adaptive word detection (blue curves) shows overall better results compared to the fixed one (pink curves), even at lower latency. 
This suggests that 
%making the comparison between 
comparing
the two strategies by using models with 
%a 
higher translation quality 
%highlights 
shows
the superiority of 
%the 
adaptive word detection at any latency regime.

%This means that using weak models to select the best performing simultaneous system and its characteristics, as the word detection strategy, can lead to a wrong decision. 
%Therefore, w
In light of these results, we conclude
that 
%\emph{i)} 
\textbf{data augmentation 
%techniques improve 
improves
%the quality of the offline-trained system 
the offline-trained system quality
without affecting latency}.
%, and \emph{ii)} \textbf{the use of high-resource systems is recommended to choose the best SimulST settings to adopt}. \mn{[MAH...]}
%correctly select the best SimulST components.
To better assess these performance gains in the simultaneous framework, in the next section we present a detailed comparison of our offline-trained Conformer with 
%the current SimulST state of the art.
the state-of-the-art SimulST architecture.
%in the following section.
%In the following, we present a comparison between our best offline-trained SimulST system, the Conformer-based architecture trained with the additional KD data, and the current state-of-the-art model.

\begin{figure*}[htb]
     \centering
     \begin{subfigure}[b]{0.49\textwidth}
         \centering
         \includegraphics[width=\textwidth]{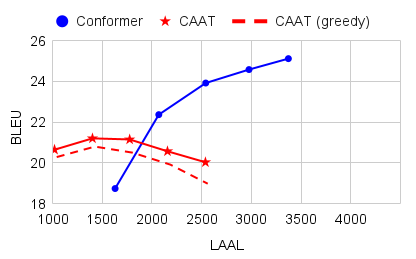}
         \includegraphics[width=\textwidth]{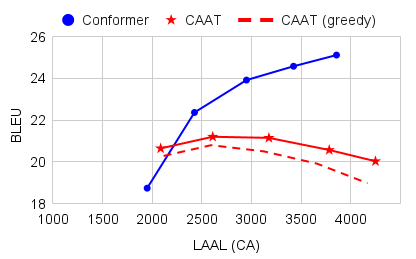}
        \caption{English $\rightarrow$ German}
     \end{subfigure}
     \hfill
     \begin{subfigure}[b]{0.49\textwidth}
         \centering
         \includegraphics[width=\textwidth]{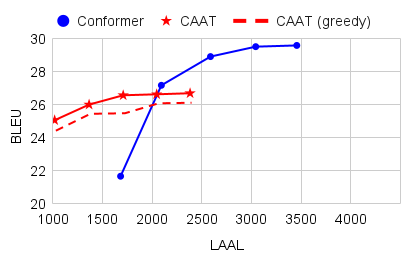}

         \includegraphics[width=\textwidth]{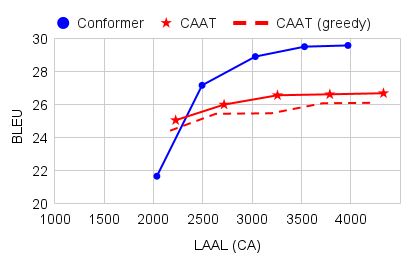}
         \caption{English $\rightarrow$ Spanish}
     \end{subfigure}
        \caption{LAAL/LAAL\textsubscript{CA}-BLEU curves of our offline-trained Conformer and state of the art (CAAT) models.}
        \label{fig:caat}
\end{figure*}

\section{Comparison with the state of the art}
\label{sec:caat}
So far, we discovered that scaling to better performing architectures and more data further improves the simultaneous 
%performance 
results
of offline-trained models. But how good is their performance compared to the 
%current state of the art?
state of the art in SimulST?
% adopting a Conformer-based architecture and data augmentation techniques further improves the performance of our offline-trained SimulST system.
% found the offline-trained system
% %, to which the segmentation strategy is applied at inference time, 
% to be the best training strategy for the SimulST models based on \emph{wait-k} policy. We also discovered that adopting the Conformer-based architecture and augmenting the training data via KD further improves the simultaneous performance other than the offline one. 
%Additionally, the adaptive proved to be the best word detection strategy when considering high-resource systems. 
%But what is the performance of our offline-trained Conformer compared to the current state of the art?
To 
%find the answer,
answer this question,
we compare our best system, the offline-trained Conformer with adaptive word detection, with the Cross Attention Augmented Transducer \citep{liu-etal-2021-cross} 
%-- or CAAT --
-- CAAT --
%that is the current SimulST state of the art.
%\mn{[DA COSA SI CAPISCE CHE CAAT E' LO STATO DELL'ARTE? IWSLT 2022 CI DICE QUALCOSA A RIGUARDO?]}.
used by the winning submissions at IWSLT 2021 \cite{iwslt_2021} and 2022 \cite{iwslt_2022}.
Inspired by the Recurrent Neural Network Transducer by \citet{graves2012sequence}, 
CAAT is made of three Transformer stacks: the encoder, the predictor, and the joiner. These three elements are jointly trained in simultaneous to optimize the quality of the translations while 
%taking 
%the 
keeping
latency under control.
%by considering all possible sequences of \texttt{READ-WRITE} actions, 
%also adopting a latency loss to ensure the system latency is under control.
%, making CAAT a SimulST model with learned decision policy.
%. These last characteristics make CAAT a SimulST model with a learned decision policy.
%also using a novel latency loss ensuring the latency of the CAAT system to be under control.

For training and testing the CAAT architecture, we use the code published by the authors and adopt the same hyper-parameters of their paper.
%The target vocabulary is the same used for the Conformer-based architectures but, since the CAAT model is implemented using only SentencePiece, we use a SentencePiece vocabulary for the source side with the same dimension of 5000 of our vocabulary.
As the performance of the CAAT model is sensitive to sequence-level KD (\citealt{liu-etal-2021-cross} show a 2 BLEU degradation without it),
%To achieve better performance, the CAAT model has to be trained using  KD data,\footnote{The CAAT authors show a quality degradation of about 2 BLEU points without the KD data.} as we have done in Section \ref{subsec:KD}. %This would also allow us to make a fair comparison between the two models.
%Thus, to make a fair comparison, we choose to compare the offline-trained Conformer model using the same data settings of CAAT, that is with the addition of the KD data as in Section \ref{subsec:KD}.
we compare it with the offline-trained Conformer model using the same data settings -- see Section \ref{subsec:KD}.
We report the CAAT results obtained by adopting both the greedy search used in our SimulST settings and the beam search used by 
%\citealt{liu-etal-2021-cross}.
\mn{\citet{liu-etal-2021-cross}.}
%used by the authors.
%The $k_{test}$ used for testing are the ones used in Section \ref{sec:waitk}.
As suggested by \citet{ma-etal-2020-simulmt}, we also compute the Computational Aware (CA) version of the LAAL metric (LAAL\textsubscript{CA}), which is defined as the time elapsed from the beginning of the generation process to the prediction of the partial
%\mn{complete} 
target.\footnote{Given that LAAL\textsubscript{CA} depends on the computation time, we perform all the generations on 
%a GPU K-80
one NVIDIA Tesla K-80 GPU
%\mn{the same NVIDIA Tesla K-80 GPU}
and provide the results by averaging over 3 runs. However, we notice a very small variance among the runs (in the order of $10ms$), suggesting that averaging is not necessary to provide sound results.} 
Since LAAL\textsubscript{CA} represents the real wall-clock elapsed time experienced by the user,
%Since it considers the total elapsed time to get the partial output translation,
it gives a more reliable evaluation of the SimulST performance in a 
%real live 
%\mn{real-life}
real-time
scenario.
For the sake of completeness, we also report the 
results 
%in terms 
of 
Average Lagging \citep{ma-etal-2020-simuleval} in Appendix \ref{sec:AL}. 
%\mn{[NELL'APPENDICE NON PRESENTEREI SOLO I PLOT MA LI ACCOMPAGNEREI CON UN COMMENTO, POSSIBILMENTE CHE TIRI ACQUA AL TUO MULINO!]}

%We also adopt the Computational Aware version of the AL metric (AL\textsubscript{CA}) proposed by \citet{ma-etal-2020-simulmt} which considers the computational time in the evaluation as the elapsed time in $ms$ needed for the generation.

We present the 
%comparison with the state of the art
comparison
in Figure \ref{fig:caat}.
%\paragraph{Non Computational Aware Results.} 
From the LAAL-BLEU curves, we 
%can see 
see
that,
%The AL-BLEU curves show that, 
at
%a 
low latency regime, the CAAT model (in solid red) outperforms our offline-trained Conformer model (in solid blue) by 2 BLEU
%\mn{points} 
on en-de and 4 BLEU
%\mn{points} 
on en-es. 
However, moving to medium-high latency regime, the Conformer significantly outperforms CAAT, reaching gains of 4 BLEU 
%\mn{points} 
on en-de and 2 
BLEU
%\mn{points} 
on en-es. 
We can also notice a degradation of the CAAT en-de translation quality that is caused by an under-generation problem at higher 
%latency, for which detailed results are presented in 
%latency, which we show more in detail in Appendix \ref{sec:overgen}.
latency, for which we give details in
Appendix \ref{sec:overgen}.
%\mn{[STESSO COMMENTO FATTO PRIMA: NELL'APPENDICE NON PRESENTEREI SOLO LA TABELLA MA L'ACCOMPAGNEREI CON UN COMMENTO, POSSIBILMENTE CHE TIRI ACQUA AL TUO MULINO! INOLTRE, QUI PARLI DI ``under-generation'' MENTRE IL TITOLO DELL'APPENDICE B E' ``Over-generation statistics''; ROBA DA POCO, MA E' MEGLIO ESSERE CONSISTENTI.]}

%As already noticed in Section \ref{subsec:adaptive}, the quality of the \emph{conformer adaptive} translation greatly increases as the latency increases. A similar behaviour is shown by the Conformer with fixed segmentation (\emph{conformer fixed}) that, however, has an overall lower quality compared to the \emph{conformer adaptive}.
%\paragraph{Computational Aware Results.} 
When it comes to 
%the AL\textsubscript{CA}-BLEU curves, 
LAAL\textsubscript{CA}-BLEU,
the scenario changes, bringing CAAT curves much closer to those of Conformer. 
%The latency difference between CAAT and Conformer seems to be only theoretical while being near on par in real-time. 
The state of the art still outperforms the Conformer at lower latency but in this case, waiting about $100/200ms$ more, the Conformer performance starts to improve consistently.
%In light of these results, we can conclude that our offline-trained system obtains higher performance at medium-high latency compared to the state of the art.
%, with all the advantages of having a single architecture to serve both the offline and simultaneous scenario.
%\paragraph{Non Computational vs Computational Aware.}
%%%%%%%%%%%%%%%%
%To conclude, we can say that \textbf{our offline-trained Conformer achieves better performance at medium and high latency compared to the state of the art}.

Comparing the LAAL- and LAAL\textsubscript{CA}-BLEU curves, we see that our offline-trained system is more coherent between computational and non-computational
aware metrics: while 
%the 
Conformer has a computational overhead of $400/500ms$, 
%the 
CAAT requires $1400/1500ms$ more than its ideal LAAL.
The CAAT greedy curves (dotted red) show only a little improvement in latency compared to the beam search (solid red), suggesting that its higher computational cost 
%is not attributed to the generation strategy but depends on other factors as
does not depend on 
the generation strategy but on other factors like its complex and more 
%computational
computationally
expensive architecture.
All in all,
we can say  
%that \textbf{our offline-trained Conformer achieves better performance at medium and high latency compared to the state of the art}, especially when the computational aware metrics are considered for the evaluation.
that, \textbf{compared to the state of the art in SimulST, the lower performance of our offline-trained Conformer at
%the 
low latency regime is balanced by consistently higher BLEU scores at medium and high latency.}
%and \emph{ii)} \textbf{the use of computational aware latency metrics is recommended to provide a correct and reliable evaluation of the SimulST systems performance in real-time}.

%at inference time.
%with the limitations of having a complex architecture 

%This highlights the poor computational speed of more complex systems such as CAAT
%, highlighting the poor computational speed of more complex systems such as CAAT. This points out that the computational time is a key factor that has to be considered to correctly and reliably evaluate a SimulST system in real-time.

\section{Conclusions}
To reduce the potentially large amount of experiments usually performed to build SimulST models, we explored the use of 
%an
a single
offline-trained model to serve both the offline and simultaneous tasks.
Through comparison with 
%the 
native SimulST systems, we showed that 
%the
our
offline-trained model can be 
%effectively 
successfully
used in real-time, achieving comparable or even better results. 
To further enhance its performance, we 
%also
investigated the adoption of consolidated techniques and emerging architectures from 
%the 
offline research, showing consistent improvements
%in the simultaneous scenario and suggesting the need for high-resource models to make more sound decisions about the best SimulST settings to adopt.
also in the simultaneous scenario.
%%The final comparison with the state of the art pointed out the limitations of considering non computational aware metrics for latency, potentially causing a misleading evaluation of the SimulST system performance in real-time.
The 
%results achieved by the offline-trained model
benefits of offline training
indicate the potential of applying this method 
%to different offline architectures
without
the need for any additional training or 
%adaptation to perform the SimulST task.
adaptation.
% \sara{This inevitably reduces the carbon footprint of the ST trainings since not only one model has to be trained to perform the simultaneous task but also to perform the offline one.
% Moreover, it also allows us to re-use already trained systems with no additional environmental impact, representing an important step for AI towards energy saving. }
Besides facilitating system deployment, 
another important 
%%side-effect
advantage of
%\mg{the benefits of offline training include the need to build and mantain a single system, and}
building and reusing one single model to rule both tasks is
the drastic reduction of the carbon footprint of ST training (by a factor of
%8
9 in our evaluation setting). This
%reduction of the environmental impact 
represents an important step in response to rising concerns about the AI energy consumption and environmental impact 
%and its un
toward more
sustainable development.
%, especially at a medium-high latency regime. Except at lower latency, this allows the model to achieve better results compared to the state of the art, both considering AL but especially AL\textsubscript{CA}.

% Concerning the SimulST evaluation,
% the differences between the 
% %AL and AL\textsubscript{CA}-BLEU curves 
% non computational and computational aware latencies
% suggest that including or not the computational time in the metrics calculation heavily 
% %impacts 
% influences
% the systems comparison.
% %the evaluation of these systems. 
As regards SimulST evaluation, the differences between results computed with 
non-computationally and computationally aware latency metrics suggest that 
%considering the computational time when evaluating performance
including computational time in the measurements
heavily influences the outcomes of system comparisons.
% In particular, the computational unaware AL metric
% depicts a big difference in terms of latency between the offline-trained and the state of the art models, which becomes smaller when its computational aware version is considered.
In our particular case, the  differences in latency between the offline-trained models and the state of the art observed in terms of the 
%computational unaware
non-computationally aware
LAAL metric become smaller when considering its computational aware version.
Although lower latency is theoretically reached by the state of the art 
%but conflicts with a 
CAAT model, this comes at the cost of a
more complex and 
%computational
computationally
expensive architecture
%, showing 
that shows
its limitations at inference time.
%when a more reliable evaluation setting is considered.
%We therefore recommend 
We therefore invite
the SimulST community to use computationally aware metrics for 
%systems evaluation and to refer
more sound evaluations, referring to ideal metrics only in the absence of similar testing assets, as machines with comparable computational power.

% Avoiding high computational costs will remain the focus of our following research and future works will be mainly devoted to improve the decision policy, exploring also alternatives to the \emph{wait-k}, 
% %with the aim of reducing 
% for reducing latency at low latency regime.
%the generation strategy by working both on the quality but also on the latency side. Analyzing and developing more performing decision policies, especially aimed at reducing latency, will be the main focus of our future research.

\section{Limitations}
% \mn{Two limitations of this work concern the breadth of the 
% %research
% exploration
% from the point of view of the technological solutions 
% %explored
% considered
% and the covered languages.}

%Despite the 
Although it
% exploits
relies on 
a simpler architecture and generation strategy compared to the state-of-the-art 
%system, the offline model
in SimulST, our offline-trained model
exhibits a high translation quality in real-time, 
%enabling it
which allows it
to achieve better results at medium and high latency regimes. However,  
%it is far from being competitive with the state of the art. 
a performance gap of 2-3 BLEU points is still observed at low latency regime.
%The cause of this performance gap 
This
can be attributed to the use of a simple policy such as \emph{wait-k}. 
Being the most popular and widely adopted one, we chose to focus on this policy 
%in conducting our analysis but more 
to conduct our analysis. 
% More }
% complicated and potentially 
% %more
% \mn{better}
% performing 
% %ones should be studied or proposed.
% \mn{solutions 
% %present in the literature
% will be investigated in future work}
%However,
Notwithstanding,
investigating better performing solutions to boost performance at low latency and close the gap is still necessary, and definitely among our future work priorities. 

%are present in the literature and could have been investigated.

%\mg{This work has focused on Transformer-like architectures, the actual state-of-the-art model in ST. Our findings might not apply for completely different architectures that may be proposed in the future.
%
%
%
%Finally, the experiments presented in the paper are limited to latin and germanic languages, so their extension to different languages, such as asiatic ones, should be verified.
Also, the experiments presented in the paper are limited to 
%latin and germanic languages.
two target languages, which were selected as representatives of those having similar and different word ordering with respect to the English source speech.
%(having respectively similar and different word ordering with respect to English).
Although this choice allowed us to reliably test our hypotheses in diverse conditions,
verifying our findings on a wider set of languages is 
%the other
another
natural evolution of this research.

\section*{Acknowledgments}
This work has been supported by the project Smarter Interpreting\footnote{\url{https://smarter-interpreting.eu/}} financed by CDTI Neotec funds, by the ISCRAB project DireSTI granted by CINECA, and by the project “AI@TN” funded by the Autonomous Province of Trento.
%This work has been carried out as part of the project Smarter Interpreting (\url{https://kunveno.digital/}) financed by CDTI Neotec funds. 
%be used to improve the offline-trained system performance.
% applied to the offline system. 
%highlighting that there is room for improvement 
% We acknowledge that the portability of the offline systems to the simultaneous scenario has been only analyzed for Transformer-like architectures, the actual state-of-the-art model in many research fields of artificial intelligence.
% The rapid evolution of the models, that become more and more performing every year, might affect our findings that might not be verified for completely different architectures. 
% Suffice it to say that when we switched from Transformer to Conformer, which is only a modified version of Transformer preserving its main characteristics, and when we improved its quality by applying KD, we found the superiority of one word detection strategy over the other, a result that had not previously emerged.
% % with a less performing model. 
% This is already the demonstration that no conclusions can be drawn definitively and for every type of present and future architectures and that 
% similar works could be required on new technologies to confirm our results.
%works of this type will have to be done on new technologies to exclude any change.

%Moreover, the experiments present in the paper involve latin languages while not considering others as asiatic languages. 

\bibliography{anthology,custom}
\bibliographystyle{acl_natbib}
%\newpage
\appendix

\section{Models Architecture}
\subsection{Transformer}
\label{subsec:transformer}
The models used in Section \ref{sec:offline-results} are based on a 12 encoder and 6 decoder layers of Transformer \citep{transformer} architecture. The embedding dimension is set to 256, the number of attention heads to 4 and the feed-forward embedding dimension to 2048, both in the encoder and in the decoder. The number of parameters is 
%32,415,081.
$\sim32.4$M
We use Fairseq \citep{ott2019fairseq} library for all the trainings. The \emph{wait-k} with fixed word detection strategy was already present in the Fairseq library, while we implemented the adaptive one.
%\mg{BISOGNA DIRE CHE RILASCIAMO IL CODICE UPON PAPER ACCEPTANCE}

We use the hyper-parameters of \citep{ma-etal-2020-simulmt} for all the trainings of the Transformer-based model. 
We use a unigram SentencePiece model \citep{kudo2018sentencepiece} for the target language vocabulary of size 8,000 \cite{di-gangi-etal-2020-target}. For the source language vocabulary of size 5,000 we use a BPE SubwordNMT model \citep{sennrich2016neural} with Moses tokenizer \citep{koehn-etal-2007-moses}. The reason for which we used
%the 
%Moses
SubwordNMT
instead of SentencePiece
%tokenizer is
lies in the strategy used for determining the end of a word, which is crucial for simultaneous inference.
While SentencePiece uses the character \enquote{$\_$} at the beginning of a new word, SubwordNMT appends \enquote{@@} to any token that does not represent the end of a word.
%uses \enquote{@@} to indicate the middle of a word. 
Thus, SentencePiece units require
%requires to generate
the generation of the first token of the next word to determine if the
%actual one is finished
current word is over while SubwordNMT
units do not.
%not.
For instance, 
the sentence \enquote{this is a phrase}, is encoded into SentencePiece units as \enquote{$\_$th is $\_$is $\_$a $\_$ph rase}. As such, to determine if \enquote{$\_$th is} is a complete word,
%in SentencePiece we have \enquote{$\_$th is $\_$is $\_$a $\_$ph rase}, to determine if \enquote{$\_$this} is finished, 
we have to wait for the next word with the \enquote{$\_$} character at the beginning, that is \enquote{$\_$is}.
Instead, with SubwordNMT we have \enquote{th@@ is is a ph@@rase }, and we 
%have not to wait
do not need to receive \enquote{is} to determine that \enquote{th@@ is} is finished.
%For this reason, we choose to adopt SubwordNMT BPE for the source vocabulary.

We select the best checkpoint based on the loss and early stop the training if the loss did not improve for 10 epochs. We trained the system for 100 epochs at maximum. At the end of the training, we make the average of the 7 checkpoints around the best one.

For the inference part, we use the SimulEval tool \citep{ma-etal-2020-simuleval} as in \citep{ma-etal-2020-simulmt} with the additional \texttt{force\_finish} tag
%, already present in SimulEval, that
that forces the model to generate text until the source speech has been completely ingested, i.e. to ignore the end of sentence token if predicted before the end of an utterance.
%, that is to ignore the end of sentence token if predicted, until the source speech has been completely ingested. 
In case of \emph{wait-k} with adaptive word detection, we also force the model to predict the successive most probable token if the end of sentence is predicted (that we called \texttt{avoid\_eos\_while\_reading}), while for the fixed we found that it degrades the performance.
The detection is taken every average word duration, that is every $280ms$, as estimated by \citet{ma-etal-2020-simulmt} in the MuST-C dataset.

\subsection{Conformer}
\label{subsec:conformer}
For the Conformer model, we build an architecture similar to \citet{inaguma2021non}, we use 12 Conformer encoder layers and 6 Transformer decoder layers. The number of parameters is $\sim35.7$M.
%around 35,698,537.  
%\mg{O IS AROUND 36M, O DIRE AROUND UN NUMERO PRECISO E' STRANO. PRIMA C'ERA UN NUMERO PRECISO PER IL TRASFORMER. ORA DATO CHE IMMAGINO CHE GIA' SOLO TRA TEDESCO E SPAGNOLO NON SIANO PROPRIO UGUALI, IO METTEREI DAPPERTUTTO $\simX$M} 
We use the same embedding dimension of our Transformer-based architecture, 4 attention encoder heads and 8 attention decoder heads. 
For the Conformer Feed-Forward layer, Attention layer, and Convolution layer, we use 0.1 as dropout. We use a kernel size of 31 for the point- and depth-wise convolutions of the Convolution layer. 
The vocabularies are the same of the Transformer-based, as well as the selection of the checkpoint. At inference time, the \texttt{force\_finish} tag is used with the \texttt{avoid\_eos\_while\_reading} for both the word detection strategies.

\subsection{Machine Translation}
\label{subsec:mt}
The MT model used to generate the target for the KD was trained on OPUS datasets \citep{tiedemann-2016-opus}. It is a plain Transformer with 16 attention heads and 1024 features in encoder/decoder embeddings, resulting in 212M parameters. The English$\rightarrow$German MT scores 32.1 BLEU and the English$\rightarrow$Spanish MT scores 35.8 BLEU on MuST-C tst-COMMON.
%\mg{MMMH, SICURA CHE ERANO SU TUTTO OPUS? MI SEMBRANO BASSI COME SCORE PER ESSERE SU OPUS, NON ERANO SU WMT?}

\section{Under-generation Statistics}

In Section\ref{sec:caat}, while discussing
%From 
the en-de curves of Figure \ref{fig:caat}, we 
%noticed
highlighted
a performance degradation of CAAT at higher latency regimes. In fact, during our experiments we observed that CAAT tends to generate
%incurs in an under-generation problem as long as the values of $k$ increase. 
shorter sentences as the value of $k$ increases.
This 
%phenomenon is shown
behaviour becomes 
%visible
apparent
in Table \ref{tab:overgen_stat}, where we report the word length difference between the generated hypotheses and the corresponding references. 
For en-de, CAAT exhibits 
%an under-generative tendency
a strong tendency to under-generate
(indicated by negative values) at high latency and this is presumably the reason why we observed the BLEU drop.

%while our offline Conformer shows the opposite trend since it under-generates more at lower frequency besides 

\label{sec:overgen}
\begin{table}[hbt]
    \centering
    \setlength{\tabcolsep}{4pt}
    \begin{tabular}{c|c|c|c|c|c}
        \hline
        \multicolumn{6}{c}{\textbf{English$\rightarrow$German}} \\
        \hline
        \texttt{Model} & \texttt{k=3} & \texttt{k=5} & \texttt{k=7} & \texttt{k=9} & \texttt{k=11} \\
        %\multicolumn{2}{c|}{\textbf{En-De}} & \multicolumn{2}{c}{\textbf{En-Es}} \\
        %\cline{2-5}
         \hline
         Conformer & -1 & -0.94 & -0.93 & -0.77 & -0.63 \\
         CAAT & 0.47 & -0.3 & -0.79 & -1.26 & -1.55 \\
         \hline
        \multicolumn{6}{c}{\textbf{English$\rightarrow$Spanish}} \\
        \hline
        \texttt{Model} & \texttt{k=3} & \texttt{k=5} & \texttt{k=7} & \texttt{k=9} & \texttt{k=11} \\
        %\multicolumn{2}{c|}{\textbf{En-De}} & \multicolumn{2}{c}{\textbf{En-Es}} \\
        %\cline{2-5}
         \hline
         Conformer & 0.48 & 0.49 & 0.53 & 0.74 & 0.80 \\
         CAAT & 1.57 & 0.96 & 0.61 & 0.35 & 0.18 \\
        \hline
    \end{tabular}
    \caption{Average word length difference w.r.t. the reference. Positive values indicate exceeding words, negative values indicate missing words.}
    \label{tab:overgen_stat}
    \end{table}

\newpage
\section{Average Lagging}
\label{sec:AL}

\begin{figure}[!htb]
     \centering
     \begin{subfigure}[b]{0.485\textwidth}
         \centering
         \includegraphics[width=\textwidth]{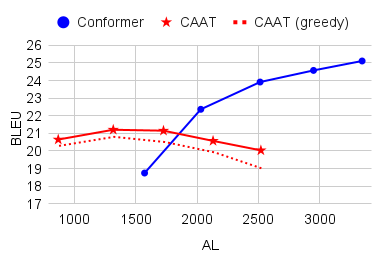}
         \includegraphics[width=\textwidth]{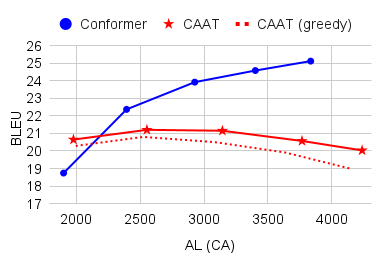}
        \caption{English $\rightarrow$ German}
    %  \end{subfigure}
    %  \hfill
    %  \begin{subfigure}[b]{0.37\textwidth}
         \centering
         \includegraphics[width=\textwidth]{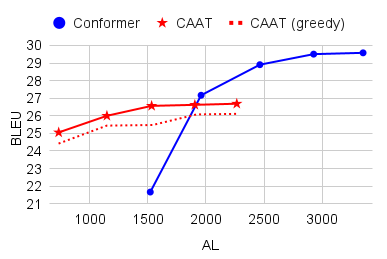}
         \includegraphics[width=\textwidth]{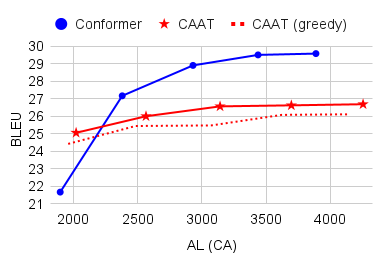}
         \caption{English $\rightarrow$ Spanish}
     \end{subfigure}
        \caption{AL/AL\textsubscript{CA}-BLEU curves of our offline-trained Conformer and 
        CAAT models.
        }
        \label{fig:dal}
\end{figure}

\end{document}